\documentclass[10pt,twocolumn,letterpaper]{article}

\usepackage{cvpr}              

\usepackage[dvipsnames]{xcolor}

\definecolor{cvprblue}{rgb}{0.21,0.49,0.74}
\usepackage[pagebackref,breaklinks,colorlinks,citecolor=cvprblue]{hyperref}

\def\paperID{5722} 
\def\confName{CVPR}
\def\confYear{2024}

\title{Object Dynamics Modeling with Hierarchical Point Cloud-based Representations}

\author{Chanho Kim \\
Oregon State University \\
{\tt\small kimchanh@oregonstate.edu}
\and
Li Fuxin\\
Oregon State University\\
{\tt\small lif@oregonstate.edu}
}

\begin{document}
\maketitle
\begin{abstract}
Modeling object dynamics with a neural network is an important problem with numerous applications. Most recent work has been based on graph neural networks. However, physics happens in 3D space, where geometric information potentially plays an important role in modeling physical phenomena. In this work, we propose a novel U-net architecture based on continuous point convolution which naturally embeds information from 3D coordinates and allows for multi-scale feature representations with established downsampling and upsampling procedures. Bottleneck layers in the downsampled point clouds lead to better long-range interaction modeling. Besides, the flexibility of point convolutions allows our approach to generalize to sparsely sampled points from mesh vertices and dynamically generate features on important interaction points on mesh faces. Experimental results demonstrate that our approach significantly improves the state-of-the-art, especially in scenarios that require accurate gravity or collision reasoning. 
\end{abstract}    
\section{Introduction}
\label{sec:intro}

\begin{figure*}[ht]
  \includegraphics[width=\textwidth]{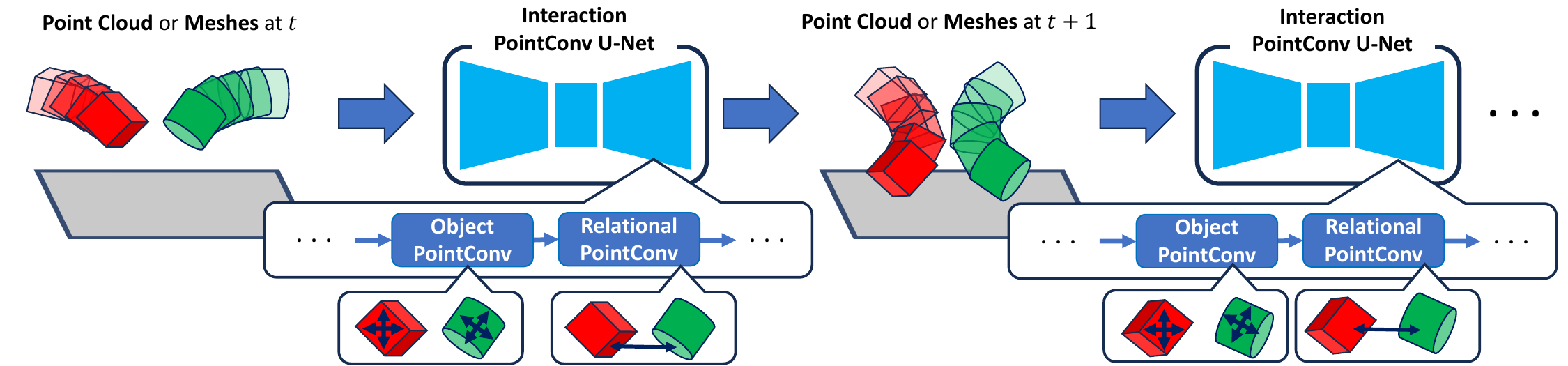}
  \caption{We propose a point-based convolutional neural network that is capable of learning object dynamics. Two different types of convolution operations, Object PointConv and Relational PointConv, are utilized alternatively to model force propagation within the same object and across different objects, respectively. A U-Net architecture encodes the point cloud into a smaller point cloud to capture long-term interactions and then decodes back to the original point cloud to make predictions.  Point-based continuous convolution allows the proposed model to be compatible with both point cloud and mesh inputs with minor modifications.} \label{general_approach}
\end{figure*}

Comprehending physics constitutes a fundamental facet of common sense knowledge. Humans naturally acquire this understanding early in life, predicting the outcome of collisions and nonlinear motions without the need to understand how to solve partial differential equations. Such capabilities of fast and intuitive physical predictions would also be crucial for enabling robots to plan actions effectively. For instance, when stacking boxes, a robot needs to understand how to arrange them to prevent the boxes from collapsing. Similarly, when a robot manipulates objects on a table to achieve a goal, it must comprehend the configuration of objects resulting from its actions. While traditional physics simulators \cite{mujoco,bullet} can be employed to grant robots such reasoning abilities, it is hard for them to generalize to planning tasks with realistic sensory inputs.

Early attempts for developing a learning-based approach for simulation aimed to learn object dynamics directly in the image pixel space \cite{NIPS2017_8cbd005a, pvoe_benchmark, qi2021learning, int_physics}, which have difficulty reasoning through complex object interactions and collisions. Recently, more promising approaches have been proposed, aiming to directly learn object dynamics in 3D space. The most popular approaches in this direction have utilized Graph Neural Networks (GNNs) \cite{gnn, msg_passing}. These models represent objects using dense particles or meshes in three-dimensional space, constructing graphs with nodes corresponding to these particles or mesh vertices, and edges linking nearby particles or representing mesh edges. Message passing networks are then learned to simulate the multi-step propagation of forces over graph nodes.

Early GNNs embed all the relational information into edge weights. However, people have increasingly realized that physics happens in the three-dimensional world and that ignoring geometry is not ideal, not to mention there is also gravity, which strongly depends on the relationship in the $z$ (vertical) direction in 3D space. Hence, later GNNs for physics and collision modeling have increasingly started to embed Euclidean space coordinates as part of the features~\cite{li2018learning,han2022learning}. More recently, as people have started to discover the importance of long-range interaction modeling, U-Net-like structures have also emerged in the form of multi-scale GNNs, which employ different types of message passing between nodes at different levels \cite{cao2022bi,grigorev2023hood,fortunato2022multiscale}. 

Rather than forcing GNNs to embed themselves into the 3D space, an alternative would be to directly employ a network that has already been designed for unordered 3D point clouds. Such networks have undergone significant improvements in the past few years with impressive capabilities to solve complex real-world recognition tasks~\cite{Schult2023mask3d,rukhovich2023tr3d,rukhovich2022fcaf3d,wu2022point}. They have also received many engineering updates to be able to handle hundreds of thousands of points as input simultaneously. Point-based approaches naturally employ the $xyz$ coordinates, and their neighborhood structure based on Euclidean distances is natural for modeling collisions in the 3D world. Besides, the spatial topology can still be modified by changing the neighborhood structure, similar to that of GNNs. Some continuously defined point-based approaches, such as PointConv~\cite{wu2019pointconv}, additionally allow interpolation to generate features at any point in the 3D space even if there was no previous feature presence. This flexibility can potentially offer a better solution for hierarchical feature modeling than graph-based approaches with between-level message passing, which would require pre-computed features at each node to process messages consistently.

Given the convergence of graph- and point-based approaches, we believe that a study that applies point-based approaches in the task of learning object dynamics is timely so that future work from both fields can borrow from each other. There is prior work that uses point-based continuous convolution for learning fluid simulation \cite{Ummenhofer2020Lagrangian}, but the models presented there were neither deep nor large-scale. Besides, it cannot be directly applied to simulating physics in scenes with a variable number of objects.

In this work, we introduce a novel point-based continuous convolution network designed to model the collision dynamics of multiple objects composed of dense 3D points. We propose novel network designs that facilitate object-centric relation reasoning on point clouds and build a U-Net architecture that learns hierarchical feature representations of the scene at different physical scales as shown in Fig. \ref{general_approach}. We demonstrate that our proposed model can effectively learn object dynamics from dense point clouds. Besides, we demonstrate an approach on a sparse point cloud defined on mesh vertices. Especially, we show how the interpolation capability of point convolutions enables the generation of point features at important locations on mesh surfaces, which helps the propagation of information between interacting objects.

In summary, we make the following contributions:
\begin{itemize}
    \item We propose point cloud convolution layers specific to learning object dynamics and assemble a U-Net structure with those layers for hierarchical modeling of object dynamics. 
    \item We extend a point cloud convolution layer for mesh collision reasoning, effectively computing collision effects between mesh faces. This is achieved by generating the features of the interaction points on the mesh faces and propagating them to the corresponding vertices.
    \item Experiments show that our approach significantly improves over state-of-the-art GNN methods.
\end{itemize}
\section{Related Work}
\label{sec:related_work}

In this section, we focus on reviewing existing literature on learning object dynamics from 3D data directly.

\subsection{GNN-based Models}

GNNs have been the predominant tool when it comes to learning object dynamics in 3D space. Models have been proposed for both learning particle-based \cite{li2018learning,han2022learning} and mesh-based simulation \cite{pfaff2021learning, sanchez2020learning, allen23a} using similar GNN frameworks. These approaches rely on message passing layers to update particle or mesh features on a graph. To promote faster information propagation, these approaches often have an additional hierarchy with sparser graph nodes (corresponding to multiple cluster centers found through k-means clustering \cite{li2018learning} or an object graph node per object \cite{allen2022learning}) connected to dense graph nodes defined in the observation space. Variants of these models have also been proposed to further improve performance through rotation-invariant features \cite{han2022learning} or a new type of graph node that explicitly encodes face-to-face interaction of the mesh \cite{allen2022learning}. In this work, we propose a continuous convolution-based alternative to these GNN-based approaches, which enables hierarchical feature learning efficiently. We do not touch upon rotation equivariance in this work, but similar approaches to incorporate such equivariance exist in point cloud networks as well~\cite{zhang2019rotation,xu2022unified,li2023improving} and can be added to our work. 

\subsection{Continuous Convolution-based Models}

A continuous convolution-based learned simulator has been proposed for fluid simulation \cite{Ummenhofer2020Lagrangian}. Although these approaches showed promising results in simulating fluids interacting with an environment, these approaches cannot be applied directly to other scenarios where multiple rigid or elastic objects interact with each other, due to the lack of specific modeling of the chemical bonds within the objects. Also, the proposed model in \cite{Ummenhofer2020Lagrangian} only had a few convolution layers unlike other modern networks \cite{wu2023pointconvformer} featuring many layers for processing dense point clouds. In contrast to these early convolution-based neural simulators, we propose a novel point-based convolutional network that is capable of learning deep feature representations of objects while modeling object interactions effectively. 

\subsection{Hierarchical Models}

Prior work has also attempted to address the limitations of GNN-based simulators with a more complicated hierarchical structure for better information propagation via both high- and low-resolution scene representations, but these networks are still shallow in the sense that it focuses on adding one additional hierarchy that is constructed in a better way than the ones based on k-means clustering or simple virtual object nodes in early work \cite{li2018learning}. A U-Net architecture \cite{physics_unet} has also been proposed for learning physics simulation with hierarchical feature representations, but relied on the 2D convolution operator, and thus its application was limited to data represented in a 2D grid. Recently, other U-Net architectures for graph networks have been proposed for meshes~\cite{cao2022bi,grigorev2023hood,fortunato2022multiscale} where coarser meshes can be used as downsampled levels, but these models do not possess the capability to work with dense point clouds. 

In point cloud networks, hierarchical models have been proposed from early on~\cite{qi2017pointnet++} and people have studied extensively different downsampling approaches, such as random downsampling, farthest point downsampling, and grid downsampling. Grid downsampling, which selects at most one point per voxel given a specific voxel size, has been demonstrated to be both computationally efficient and to improve recognition performance~\cite{thomas2019kpconv}. This could be due to the more regular density of points after grid downsampling, in comparison with other approaches.

\subsection{Other Point-based Networks}
The last few years have seen a proliferation of many point-based networks being proposed, starting from PointNet~\cite{qi2017pointnet} and PointNet++~\cite{qi2017pointnet++}. PointNet consists of MLP layers and max-pooling layers, and PointNet++ uses max-pooling locally within each point's neighborhood. Max-pooling tends to lose information about non-max features of points, hence the performance of those approaches is suboptimal. Besides PointConv and different variants of similar continuous convolution (e.g., ~\cite{boulch2020convpoint,wang2018deep}), point transformer approaches~\cite{zhao2021point,park2022fast,wu2022point} have been proposed with slightly better prediction performance. However, these approaches still lack the capability of generalizing to points without features. Another high-performance architecture is to directly utilize the 3D convolution but make it sparse by not computing the output of the convolution if the location is not occupied with a sampled point~\cite{graham20183d,choy20194d}. This works well on densely sampled point clouds, but the performance suffers if the sparsity is uneven and dynamic, hence not necessarily suitable for the object dynamics modeling task.
\section{Method}
\label{sec:method}

In this section, we introduce a new U-Net architecture that enables efficient collision dynamics modeling with a hierarchical scene representation using point-based convolutions. We begin by outlining the problem setup, describing the input and output of the model (Sec. \ref{problem_formulation}). Next, we introduce the PointConv operator~\cite{wu2019pointconv}, which serves as a basic building block for our point-based convolutional neural network (Sec. \ref{sec:pointconv}). Following this, we present novel PointConv operators designed to model interactions within and between objects (Sec.~\ref{sec:rel_pointconv}), as well as with dense and mesh-based sampling of points. Then, we present an interaction PointConv block (Sec.~\ref{sec:int_pointconv}) and use it to build an Interaction PointConv U-Net (Sec.~\ref{sec:int_unet}). Finally, we discuss training details and loss functions (Sec.~\ref{sec:training}).

\subsection{Problem Formulation} \label{problem_formulation}
Given point cloud observations at time $\{t-h+1, ..., t-1, t\}$ where $h$ is the length of the input history, our goal is to predict a future point cloud at time $t+1$. With the capability to predict the point cloud at time $t+1$, point clouds in the following frames can be obtained in an autoregressive manner. Each point $p_{i}(t)$ in point cloud observations has its position $p_{i}(t) = [x_i(t), y_i(t), z_i(t)]$ and velocity $v_i(t) = p_i(t) - p_i(t-1)$ as its states. The model takes a set of points as input where each point is associated with the velocity history $\{v_i(t-h+1), v_i(t-1),..., v_i(t)\}$ along with the most recent position $p_i(t)$ and the outputs $v_i(t+1)$ for all points. We set the velocity history length $h$ to 2, which essentially encodes information from $3$ frames. We concatenate velocities in the input history and provide them as input features for the model. Following the prior work \cite{li2018learning,han2022learning}, we also append the $z$ coordinate (i.e., coordinate along the gravity axis) of $p_i(t)$ to the input so that the model knows how far an input point is from the ground plane. Besides, the entire position vector $p_i(t)$ is used to generate convolution filter weights, as described below.

\subsection{PointConv}
\label{sec:pointconv}
Let $p_0 \in \mathbb{R}^3$ be a point where a convolution kernel is centered and $\mathcal{N}(p_0)$ be a set of neighbor points of $p_0$. An $\epsilon$-ball or $k$-nearest neighbors (kNN) neighborhood is often adopted when applying convolution to point clouds. However, the choice of neighborhood is flexible; for example, one can use a mesh vertex neighborhood if input points are mesh vertices. Let $x_i \in \mathbb{R}^{c_{\text{in}}}$ be the input features of a point $p_i$ and $y_i \in \mathbb{R}^{c_{\text{out}}}$ be the output features of $p_i$. The naive formulation of PointConv is defined as follows:
\begin{equation} \label{eq:original_pointconv}
y_0 = \sum_{p_i \in \mathcal{N}(p_0)} (W(p_i - p_0))^\top x_i
\end{equation}
where $p_0$ represents a query point, $y_0 \in \mathbb{R}^{c_{out}}$ represents the output feature vector of point $p_0$ and $W(\cdot)$ represents a matrix-valued function with output dimensionality $c_{in} \times c_{out}$. A nice property of PointConv is that a query point $p_0$ can be any arbitrary point in the 3D space even with $x_0$ undefined. This has enabled a PointConv layer that can either down-sample or up-sample input points using sparser or denser query points~\cite{wu2019pointconv,wu2023pointconvformer}.

This formulation, however, is computationally expensive as the size of the tensor for the backward pass of the weight tensor $W$ is equal to $c_{\text{out}} \times c_{\text{in}} \times k \times n$ for $n$ input points with a $k$-NN neighborhood.
In \cite{wu2019pointconv}, an efficient formulation of PointConv was proposed by redefining the filter weight $W(\cdot) = W_l h(p_i - p)$, leading to the following formulation: 

\begin{equation}
y_0 = W_l \mathrm{vec}(\sum_{p_i \in \mathcal{N}(p_0)} h(p_i - p_0) x_i ^\top)
\label{eq:efficient_pointconv}
\end{equation}
where $h(\cdot) \in \mathbb{R}^{c_{\text{mid}}}$ outputs a $c_{\text{mid}}$-dimensional vector, $\mathrm{vec}(\cdot)$ is an operator that flattens a matrix into a vector, and $W_l \in \mathbb{R}^{c_{\text{out}} \times c_{in} c_{\text{mid}}}$ represents a linear transformation. The size of the tensors becomes significantly smaller because, first, $W_l$ is learned and shared by all input points, besides, $c_{mid}$ can be set to as small as $4$ without significantly compromising the performance~\cite{wu2023pointconvformer}. 
This efficient PointConv formulation enables a point-based convolution network that can work with hundreds of thousands of points in dozens of layers and still fit in the memory of a single GPU.

Compared to GNNs, PointConv generates weights from relative coordinates and hence has a learned, multiplicative relationship between the coordinates and the features. This can be helpful in capturing relationships that are hard to obtain solely through concatenating positional embeddings like in GNNs \cite{allen23a,li2018learning}, such as the notion that effects from a point should be diminished or strengthened if the points are at a certain direction from each other. In the experiments, we show a direct ablation against graph convolution with positional embeddings, which indicates that the point convolution approach outperforms the alternative approach.

\begin{figure*}[ht]
  \includegraphics[width=\textwidth]{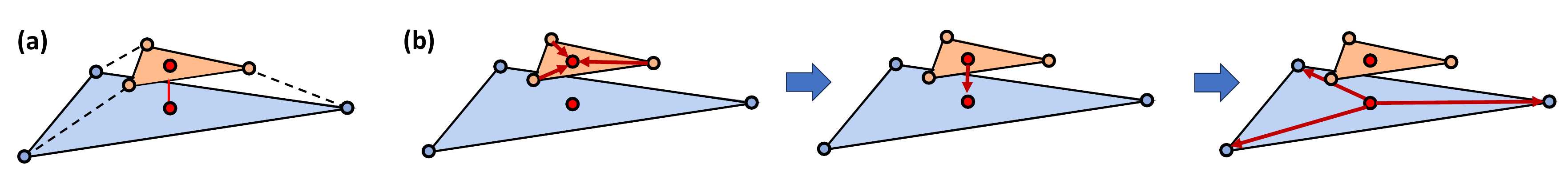}
  \caption{(a) As pointed out by \cite{allen2022learning}, given two mesh faces of different objects, a neighborhood based on mesh vertices (dotted lines) may not capture proximity between two mesh faces (red solid line), depending on the location where a collision occurs. (b) We model face-to-face collision with three PointConv layers using dynamically selected interaction points (red dots) on the surfaces.} \label{relational_pointconv}
\end{figure*}

\subsection{Learning Collision Dynamics with PointConv}
\label{sec:rel_pointconv}
When learning collision dynamics with neural networks, effects of physical force need to be propagated within objects and across different objects \cite{han2022learning}. These effects are different because particles within the same object have strong chemical bonds, whereas those forces are usually negligible between different objects. Hence, it is ideal to have a special treatment for within-object interactions. We utilize two types of PointConv operators to achieve this goal. We call the convolution operator that propagates effects within objects \textit{Object PointConv} and the other convolution operator that propagates effects across different objects \textit{Relational PointConv}. Below, we propose different modifications to the PointConv operator (Eq. (\ref{eq:efficient_pointconv})) for these layers. 

\subsubsection{Object PointConv}
As described in Sec. \ref{problem_formulation}, the input features for Object PointConv in the first layer can then be written as $x_i'(t) = f([v_i(t); v_i(t-1); z_i(t)])$, which concatenates the information and input it to a multi-layer perceptron (MLP) $f$ that outputs point-wise features from input points. Note that this does not include any information encoding a surface shape around a point $p_0$, which is important for learning collision dynamics. Hence, we additionally concatenate the relative positional embedding with the input features. The input $x_i(t)$ around $p_0(t)$ in Eq. (\ref{eq:original_pointconv}) is then defined as:
\begin{equation} \label{eq:obj_pointconv}
x_i(t) = [x_i'(t); e(p_i(t) - p_0(t))]
\end{equation} 
where $e(\cdot)$ is an MLP that takes the relative coordinates between a query point and a neighbor point as input and generates the positional embedding vector. We select the $k$ nearest neighboring points within the same object as the neighborhood for point cloud inputs and the mesh vertex neighborhood for mesh inputs. 

\subsubsection{Relational PointConv for Point Cloud}

For Relational PointConv, we would like to model point relationships with points from other objects; hence, we only search for neighbors that do not belong to the same object. Furthermore, we have noticed that including points from a faraway distance slows down learning since the network needs to learn to ignore those points, hence we filter the k-NN neighborhood $N_{\text{relation}}$ to only retain those that are within a distance $r$. This gives us a neighborhood with $\leq k$ points which we denote as $\mathcal{N}_{\text{rel}}$. In order to account for a variable number of neighbors found in the filtered neighborhood, we divide the output of the intermediate operation by the number of neighbors as follows:

\begin{equation}
y_0 = W_l \mathrm{vec}(\frac{1}{|\mathcal{N}_{\text{rel}}|} \sum_{p_i \in \mathcal{N_{\text{rel}}}(p_0)} h(p_i - p_0) x(p_i)^\top).
\end{equation}
This formulation works fine for point clouds densely sampled from object surfaces because the neighborhood always includes points from other objects when two objects collide. In the next part, we introduce a variant of Relational PointConv that takes mesh vertices as input which can be sparsely sampled from object surfaces.

\subsubsection{Relational PointConv for Mesh}

In the case of polygonal input that can be represented with a simple mesh, densely sampling from object surfaces could be wasteful~\cite{allen2022learning}. In GNNs, \cite{allen2022learning} proposed a combination of graph nodes with mesh vertex nodes and mesh face nodes, which complicates the network by requiring multiple types of interactions. Furthermore, when an object collides with another, the collision could happen at any location on a mesh face (see Fig. \ref{relational_pointconv} (a)), so using a single node to represent each face may not capture these subtleties. In this paper, we propose a novel approach to mesh-based collision modeling by utilizing the interpolation capability of PointConv to compute the features of selected points on mesh faces, that are not part of the original network input. 

Recall that a query point in PointConv can be any arbitrary point. As long as we have the input features of the points in its neighborhood $\mathcal{N}_{\text{rel}}(p)$, we can obtain the output features $y(p)$ of any $p \in \mathbb{R}^3$. In order to enable mesh face-to-face collision reasoning, we first compute distances between mesh faces of different objects and find pairs of points that are close between two nearby mesh faces (refer to the supplementary for more details). We call this pair \textit{interaction points} and use it for relational reasoning as described below.

Given two nearby mesh faces, we aim to compute the effect that one face (receiver face) receives from the other face (sender face) and update the features of the receiver face's vertices accordingly. We model this process using three PointConv layers (see Fig. \ref{relational_pointconv} (b)). The first PointConv layer computes the features of the interaction point $p_{\text{si}}$ on the sender mesh face using $p_{\text{si}}$ as a query point and mesh vertices that define the sender face as its neighborhood. 
\begin{equation}
y_{\text{si}} = W_l \mathrm{vec}(\sum_{p_i \in {\mathcal{N}_{\text{mesh-vertex}}(p{_{\text{si}}})}} (h(p_i - p_{\text{si}})) x_i^\top).
\end{equation}
The number of points in ${\mathcal{N}}_{\text{mesh-vertex}}(\cdot)$ is fixed and determined by the mesh type (e.g., 3 for a triangular mesh).

The second layer then computes the features of the interaction point on the receiver face $p_{\text{ri}}$ based on the features of interaction  points on the nearby sender faces as:
\begin{equation}
y_{\text{ri}} = W_l \mathrm{vec}(\frac{1}{|\mathcal{N}_{\text{int}} (p_{\text{ri}})|} \sum_{p_i \in \mathcal{N}_\text{int}(p_{\text{ri}})} h(p_i - p_{\text{ri}}) x_i^\top) 
\end{equation}
where $\mathcal{N}_\text{int} (p_\text{ri})$ is the filtered k-NN neighborhood that includes only interaction points from nearby sender faces. 

The last PointConv layer obtains the features of each mesh vertex $p_0$ by first propagating the features of interaction points on the receiver faces to the mesh vertex location via convolution and then applying average pooling. Here, the receiver faces are the ones that include the query point $p_0$ as one of its vertices. Then the PointConv operation can be written as:
\begin{equation}
y_0 = \frac{1}{|\mathcal{N}_{\text{int-surface}}(p_0)|} \sum_{p_i \in {\mathcal{N}_{\text{int-surface}}(p_0)}} W_l \mathrm{vec}(h(p_i - p_0) x_i^\top)
\end{equation}
where $\mathcal{N}_{\text{mesh-surface}}(p_0)$ includes interaction points from the mesh surfaces to which a mesh vertex $p_0$ belongs. 

In the first and third layers, where the features are propagated via convolution, we append the positional embedding features to the input features, as in Eq. (\ref{eq:obj_pointconv}). With these three layers, we successfully propagate information from the mesh vertices of one object to the mesh vertices of another object while taking into account the location where collision is likely to occur. 

\begin{figure*}[htb]
\centering
\includegraphics[width=0.8\linewidth]{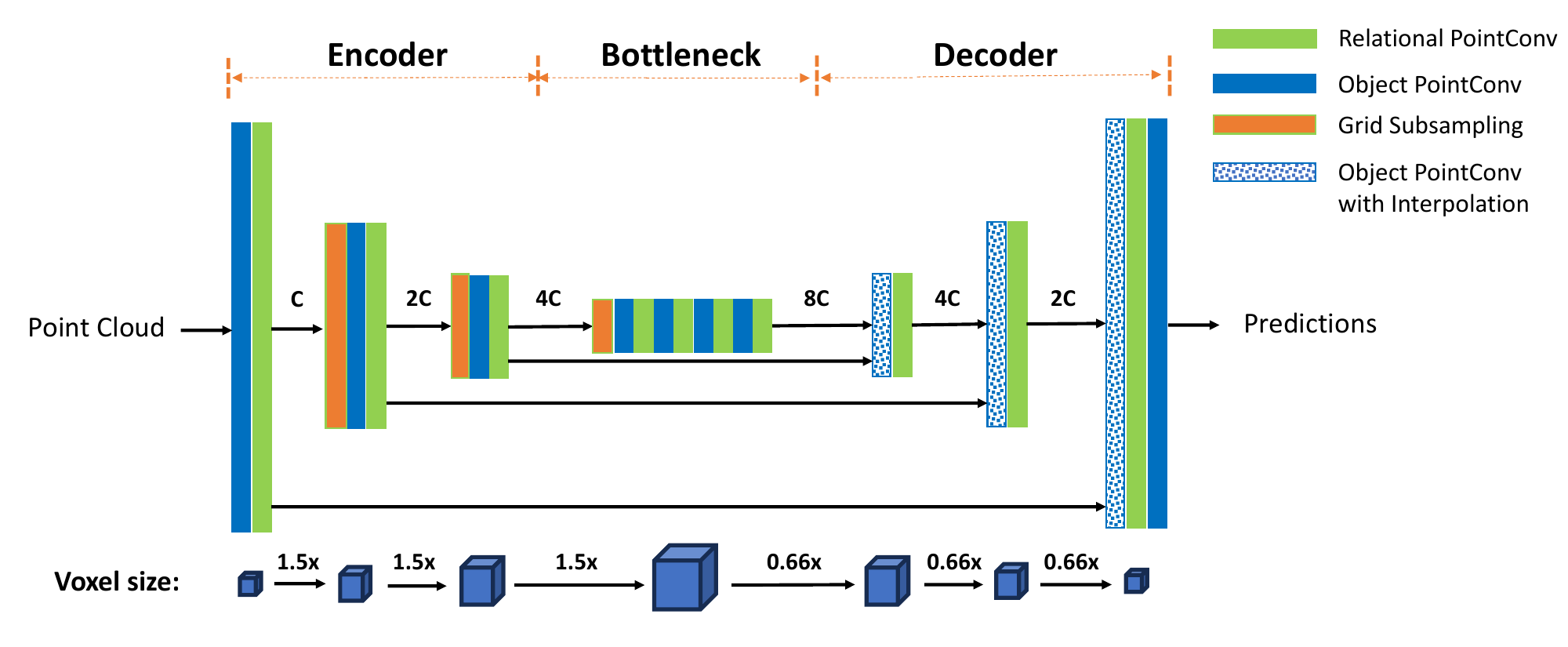}
\vskip -0.05in
\caption{The proposed U-Net architecture. The input point cloud goes through Object PointConv and Relational PointConv alternatively, with successive downsampling layers in the encoding stage. In the bottleneck layers, the voxel sizes are large and the number of points is small; hence, long-range interactions are captured with several layers. Finally, in the decoder, Object PointConv with interpolation upsamples the point clouds to the point locations at the previous level. Finally, the point cloud is upsampled back to the original size, and then pointwise velocity or acceleration is predicted. We selected $32$ as the base channel dimension $C$ for our experiments in this paper.}
\label{fig:unet}
\end{figure*}

\subsection{Interaction PointConv Block}
\label{sec:int_pointconv}
We define an interaction PointConv block as an Object PointConv layer followed by a Relational PointConv layer. The block can be set up in a way that the number of input points can change or remain the same. When point down-sampling or up-sampling is needed, we do it with sparser or denser query points in the Object PointConv layer only and maintain the spatial resolution in the Relational PointConv layer. Similar to \cite{wu2023pointconvformer}, for every PointConv layer, we have a residual connection and also utilize point-wise $1\times1$ convolutions to decrease and increase feature dimensions before and after a PointConv layer for the sake of reducing memory consumption.

\subsection{Interaction PointConv U-Net}
\label{sec:int_unet}
In order to propagate information over a long spatial range, prior work has utilized a 2-layer hierarchy, either using down-sampled points from k-means clustering as points in the upper layer \cite{li2018learning} or having a single object node connected to all the points belonging to the object~\cite{han2022learning}. In this work, we aim to learn a hierarchical feature representation of the scene with several levels in the hierarchy with a U-Net architecture shown in Fig.\ref{fig:unet}. In the encoding layers, points are downsampled successively while their locations are stored, and in the decoding layers, PointConv interpolates the features at the stored points from one level below. We follow the standard U-Net with highway connections between encoder layers and the corresponding decoder layers at the same resolution. The proposed U-Net architecture supports object interaction modeling with hierarchical object-centric scene representations from point cloud inputs. 

In this U-Net architecture, long-range force propagation across multiple objects can be cheaply modeled by adding more interaction PointConv blocks in the bottleneck while still maintaining detailed geometric features of each object learned through multiple PointConv layers. In order to build a hierarchical feature representation of the scene, we utilize grid down-sampling on the original input points, following the standard practices in the existing point cloud literature~\cite{thomas2019kpconv,wu2023pointconvformer}. 

For mesh inputs, we used a variant of the proposed U-Net architecture, integrating relational PointConv solely at the highest resolutions. This approach circumvents the need to retain mesh faces at lower resolutions, which might introduce errors, particularly when representing complicated shapes with a limited number of vertices. For further details regarding grid down-sampling for point clouds and the U-Net architecture for learning object dynamics with meshes, please refer to the supplementary. 

\subsection{Training}
\label{sec:training}
The U-Net model is capable of making point-wise predictions after the decoder layers. Following existing work \cite{allen2022learning,allen23a}, we created versions of either point-wise acceleration or velocity predictions. In the case of acceleration prediction, the network outputs $\hat{a}_p(t+1)$ for each point $p$, and the predicted position $\hat{p}({t+1})$ is then obtained by $p(t) +\hat{v}_p(t+1)$ where $\hat{v}_p({t+1}) = v_p(t) + \hat{a}_p(t+1)$. In the case of velocity prediction, the network outputs $\hat{v}_p(t+1)$ directly. We supervise the model by comparing $\hat{p}({t+1})$ against the ground truth $p({t+1})$ using the Huber loss~\cite{Huber1992} as follows:
\begin{eqnarray}
L(\hat{p}, p) & = & \frac{1}{ \sum_{i} n_i}  \sum_{i=1}^{M} \sum_{j=1}^{n_i} \text{Huber-Loss}\left(p_{ij}(t+1),  \right. \nonumber\\
&& \left. p_{ij}(t) + \hat{v}_{ij}(t_i+1))\right)
\end{eqnarray}
where $M$ is the number of point clouds in a mini-batch and $n_i$ is the number of points in the $i$th point cloud in the mini-batch. $p_{ij}$ represents the $j$th point in the $i$th point cloud.

\subsection{Inference}

We estimate a single rigid-body transformation from pointwise predictions for each rigid object to preserve the object shape throughout prediction rollouts, using the same pose fitting algorithm as in prior work \cite{li2018learning,physion,han2022learning}. For non-rigid objects, we directly use the pointwise velocity predictions.

\begin{table*} [ht!]
\centering
\scalebox{0.85}{
\begin{tabular}{lccccccccc}
\toprule
{} & Dominoes & Contain & Link & Support & Drop & Collide & Roll & Drape & Average \\
\midrule
GNS \cite{sanchez2020learning} & $78.6 \pm 0.9$    & $71.6 \pm 1.6$ & $66.7 \pm 1.5$ & $68.2 \pm 1.6$ & $65.3 \pm 1.1$ & $86.1 \pm 0.5$ & $81.3 \pm 1.8$ & $58.8 \pm 1.0$ & $74.0$ $(72.1)$   \\
DPI \cite{li2018learning} & $82.3 \pm 1.3$  & $72.3 \pm 1.8$ & $63.7 \pm 2.2$ & $64.8 \pm 2.0$ & $70.7 \pm 0.8$ & $84.4 \pm 0.7$ & $82.3 \pm 0.6$ & $53.3 \pm 0.9$  & $74.4 $ $(71.7)$ \\
SGNN~\cite{han2022learning} & $\mathbf{89.1} \pm 1.5$    & $\mathbf{78.1} \pm 1.5$ & $\mathbf{73.3} \pm 1.1$ & $71.2 \pm 0.9$ & $74.3 \pm 1.0$ & $85.3 \pm 1.1$ & $84.2 \pm 0.6$ & $60.6 \pm 0.5$  & $79.4$ $(77.0)$ \\
\midrule
Ours (vel) & $87.3 \pm 2.0$  &  $\mathbf{82.4} \pm 2.8$ & $55.1 \pm 3.5$ & $76.2 \pm 2.1$ & $\mathbf{89.8} \pm 0.3$ & $\mathbf{92.0} \pm 1.1$ & $\mathbf{86.5} \pm 0.3$  & $\mathbf{75.8} \pm 1.9$ & $\mathbf{81.3}$ $(\mathbf{80.6})$\\
\midrule
Ours (acc) & $\mathbf{90.2} \pm 1.3$ & $75.1 \pm 2.1$ &  $\mathbf{75.3} \pm 1.9$ & $\mathbf{83.6} \pm 3.5$ & $88.0 \pm 0.9$ & $\mathbf{90.9} \pm 0.8$ & $\mathbf{87.5} \pm 0.3$ & - & $\mathbf{84.4}$ $($-$)$ \\
\bottomrule
\end{tabular}
}
\caption{Physion Benchmark. We used the performance numbers reported in \cite{han2022learning} for other approaches in this table. Following \cite{han2022learning}, we also averaged our results across 3 runs. T-tests are used to determine statistical significance between pairs of approaches. The average numbers in the parentheses are the ones calculated including the Drape scenario.} \label{physion_benchmark}
\end{table*}

\begin{table*} 
\centering
\scalebox{0.9}{
\begin{tabular}{lcccccccc}
\toprule
{} & Dominoes & Contain & Link & Support & Drop & Collide & Roll & Average\\
\midrule
GNN U-Net (vel) & $87.3 \pm 2.5$  & $69.3 \pm 0.9$ & $\mathbf{74.4} \pm 2.0$ & $62.7 \pm 3.8$ & $85.5 \pm 1.4$ & $92.0 \pm 0.6$ & $86.7 \pm 0.0$  & 79.7  \\
PointConv U-Net (vel) & $87.3 \pm 2.0$ & $\mathbf{82.4} \pm 2.8$ & $55.1 \pm 3.5$ & $\mathbf{76.2} \pm 2.1$ & $\mathbf{89.8} \pm 0.3$ & $92.0 \pm 1.1$ & $86.5 \pm 0.3$ & $\mathbf{81.3}$ \\
\midrule
GNN U-Net (acc) & $88.7 \pm 1.5$  & $65.1 \pm 1.7$ & $71.8 \pm 1.9$ & $59.1 \pm 1.1$ & $84.9 \pm 0.8$ & $88.4 \pm 2.2$ & $87.3 \pm 0.3$  & 77.9 \\
PointConv U-Net (acc) & $90.2 \pm 1.3$ & $\mathbf{75.1} \pm 2.1$ & $75.3 \pm 1.9$ & $\mathbf{83.6} \pm 3.5$ & $\mathbf{88.0} \pm 0.9$ & $90.9 \pm 0.8$ & $87.5 \pm 0.3$  & $\mathbf{84.4}$ \\
\bottomrule
\end{tabular}
}
\caption{Ablation for GNN-based and PointConv-based learned simulators on the Physion dataset with the contact prediction accuracy (\%)} \label{ablation_gnn}
\end{table*}

\section{Experiments}

\begin{table}[ht!]
\centering
\scalebox{0.87}{
\begin{tabular}{lcccc}
\toprule
{} & input & face & Movi-A & Movi-C \\
\midrule
GNN U-Net (vel) & PC& - & $2.17 \pm 0.01$ & $2.58 \pm 0.03$    \\
Ours (vel) & PC & - & $2.15 \pm 0.12$ & $2.52 \pm 0.02$    \\
\midrule
Ours (vel) & M& no & $2.95 \pm 0.08$ & $2.70 \pm 0.05$     \\
Ours (vel) & M& yes & $\mathbf{2.47} \pm 0.04$ & $2.63 \pm 0.05$     \\
\midrule
GNN U-Net (acc) & PC& - &  $2.20 \pm 0.04$ & $2.55 \pm 0.08$     \\
Ours (acc) & PC & - & $\mathbf{1.87} \pm 0.01$  & $\mathbf{2.18} \pm 0.01$    \\
\midrule
Ours (acc) & M& no & $2.49 \pm 0.02$ & $2.23 \pm 0.01$     \\
Ours (acc) & M& yes & $\mathbf{1.98} \pm 0.03$ & $2.23 \pm 0.02$     \\
\bottomrule
\end{tabular}
}
\caption{Results on the Kubric dataset with the Euclidean distance as an error metric (lower is better). PC stands for point clouds, and M stands for meshes. The third column (face) indicates whether or not explicit face reasoning is enabled.} \label{kubric}
\end{table}

We conducted our experiments using the Physion \cite{physion} and Kubric \cite{greff2021kubric} datasets. The Physion benchmark provides training, validation, and testing data for seven scenarios, such as dominoes and support, involving rigid-body objects colliding as well as one scenario with non-rigid-body objects like drapes. The publicly available Kubric dataset offers training and validation data across various difficulty levels. For our experiments using Kubric, we utilized Movi-A and Movi-C videos. We excluded Movi-B as the public Movi-B data lacked object scale information necessary for generating input data for our models. In this paper, the results presented for the Physion dataset were generated using the testing data, while the results for the Kubric dataset were generated using the validation data.

As for the evaluation metrics, Physion provides the object contact prediction task, where an agent is asked to predict whether or not two query objects will contact each other based on the initial observation of the scene. Following the evaluation protocols in \cite{physion, han2022learning}, we measured the minimum distance between two objects over time while the prediction of the dynamics was made. We predict that two objects will touch if the distance comes below a threshold. This touch prediction is compared against the ground truth label. We used the contact prediction accuracy to compare against our own baselines and other recent GNN-based approaches \cite{li2018learning,han2022learning,sanchez2020learning} for Physion. For Kubric, we measured the Euclidean distance between ground truth point trajectories and predicted point trajectories at the end of prediction rollouts.

\subsection{Benchmark Results}

In this section, we compare the performance of the proposed model with other recent GNN-based approaches using the Physion benchmark. As for the Kubric benchmark, FIGNet~\cite{allen2022learning} provided evaluation results, but they generated their own Kubric sequences instead of using public Kubric sequences. As we do not have access to these Kubric sequences and the code to train FIGNet on our dataset, we could not compare against the results presented in \cite{allen2022learning}. Thus, we only present comparisons against other published work on Physion, and for Kubric, we compare against our own baselines, which include running GNNs instead of PointConv within the same U-Net architecture. 

The comparison to other GNN-based simulators is presented in Table \ref{physion_benchmark}. Both ours (vel) and ours (acc) are better than the state-of-the-art SGNN in $4$ out of $7$ scenarios involving rigid body objects, and ours (vel) is significantly better than SGNN in the non-rigid Drape scenario by more than $15\%$. The only scenario in which SGNN performs better is when compared to ours (vel) in the Link scenario. We suspect this is due to our velocity-based model being more sensitive to some label noises in the Link training data where multiple connected rigid parts are often labeled as separate rigid objects. For the other scenarios, the differences were not statistically significant. On average, ours (vel) is better than SGNN by $1.9\%$, and ours (acc) is better than SGNN by $5.0\%$ in rigid scenes. Point-based approaches are especially better than graph-based approaches in the Support, Drop, Collide, and Drape scenarios, which require accurate reasoning about gravity and object collisions. Our significantly better performance on the Drape scenario further showcases the capability of our framework to learn non-rigid-body object dynamics. 
In the case of the Drape scenario, we did not predict pointwise acceleration since the deformable motion made the acceleration unstable. Unlike SGNN/DPI, which used dense particles provided by the Physion dataset across all 8 scenarios, we used dense particles for Drape only. For our results of other scenarios with no deformable object (i.e., except for Drape in Table \ref{physion_benchmark}). we used point clouds (consisting of only surface points) extracted from meshes as input. For further details about point cloud processing, please refer to the supplementary.

Results on the Kubric dataset are shown in Table \ref{kubric}. Our approach utilizing PointConv demonstrates superior performance compared to U-Net with GNNs in acceleration prediction. Besides, we show that sampling densely on the surface yields better results than sampling sparsely on the mesh vertices (i.e., dense point cloud inputs versus mesh inputs). This is understandable as dense sampling provides much more information on the surfaces than sparse sampling. However, when we are predicting acceleration, the gap between the mesh vertices and dense samples is reduced, similar to the findings of \cite{allen23a}. Besides, we can see significant improvements when we enable collision reasoning with interaction points from the faces in the case of Movi-A where many object shapes can be accurately represented with a limited number of vertices. Explicit face-to-face collision reasoning does not appear to enhance performance in Movi-C. We speculate that in scenarios like Movi-C, where input meshes have dense vertices because of their complicated shapes, the proximity between distinct objects can still be effectively captured by measuring the distances between mesh vertices. This results in PointConv U-Net learning collision dynamics as effectively as when explicit face-to-face collision reasoning is enabled.

For additional ablation experiments on Physion, we make a comparison between the message-passing algorithm designed for GNNs and the PointConv operator within the same U-Net architecture to showcase the efficacy of point-based continuous convolution networks compared to GNNs in learning object dynamics. In Table \ref{ablation_gnn}, we compare a PointConv-based simulator against a GNN-based simulator, which is currently the most popular type of model in learning object dynamics. In order to make a fair comparison, we implemented the same message-passing steps used in \cite{li2018learning} within our U-Net architecture. Hence, the only difference would be the difference between GNN and PointConv layers. One can see that PointConv U-Net significantly outperforms GNN U-Net in the Contain and Support scenarios and also outperforms GNN U-Net in the Drop scenario. These are all scenarios that involve gravity, which shows that the multiplicative relationship between coordinates and features helps PointConv to better learn the geometric information in 3D space. 

Additionally, we include ablation results on different numbers of layers used in U-Net and on the interaction PointConv block, which creates the separation between object and relational PointConv within U-Net, in the supplementary.

\section{Conclusion}
In this paper, we propose an approach based on a U-Net structure with continuous point-based convolutions for modeling object dynamics. We extend PointConv to Object PointConv and Relational PointConv, which learn within- and between-object effects, respectively. Additionally, we propose an approach to propagate information between nearby meshes with vertex features by selecting interaction points on mesh faces dynamically and using PointConv to interpolate features on those interaction points. Experimental results demonstrate that our approach outperforms state-of-the-art graph neural network approaches, particularly on tasks involving reasoning about gravity and collisions. We hope this work can help bring the community of graph neural networks and point cloud neural networks together so that both can adopt best practices derived from the other side.
\subsection*{Acknowledgements}

This work was partially supported by the USDA AFRI award 2019-67019-29462,  DARPA award N66001-19-2-4035 and NSF award 2321851.

{
    \small
    \bibliographystyle{ieeenat_fullname}
    \bibliography{main}

\begin{thebibliography}{40}
\providecommand{\natexlab}[1]{#1}
\providecommand{\url}[1]{\texttt{#1}}
\expandafter\ifx\csname urlstyle\endcsname\relax
  \providecommand{\doi}[1]{doi: #1}\else
  \providecommand{\doi}{doi: \begingroup \urlstyle{rm}\Url}\fi

\bibitem[Allen et~al.(2023{\natexlab{a}})Allen, Guevara, Rubanova, Stachenfeld, Sanchez-Gonzalez, Battaglia, and Pfaff]{allen23a}
Kelsey~R Allen, Tatiana~Lopez Guevara, Yulia Rubanova, Kim Stachenfeld, Alvaro Sanchez-Gonzalez, Peter Battaglia, and Tobias Pfaff.
\newblock Graph network simulators can learn discontinuous, rigid contact dynamics.
\newblock In \emph{Proceedings of The 6th Conference on Robot Learning}, pages 1157--1167. PMLR, 2023{\natexlab{a}}.

\bibitem[Allen et~al.(2023{\natexlab{b}})Allen, Rubanova, Lopez-Guevara, Whitney, Sanchez-Gonzalez, Battaglia, and Pfaff]{allen2022learning}
Kelsey~R Allen, Yulia Rubanova, Tatiana Lopez-Guevara, William~F Whitney, Alvaro Sanchez-Gonzalez, Peter Battaglia, and Tobias Pfaff.
\newblock Learning rigid dynamics with face interaction graph networks.
\newblock In \emph{The Eleventh International Conference on Learning Representations}, 2023{\natexlab{b}}.

\bibitem[Battaglia et~al.(2018)Battaglia, Hamrick, Bapst, Sanchez, Zambaldi, Malinowski, Tacchetti, Raposo, Santoro, Faulkner, Gulcehre, Song, Ballard, Gilmer, Dahl, Vaswani, Allen, Nash, Langston, Dyer, Heess, Wierstra, Kohli, Botvinick, Vinyals, Li, and Pascanu]{gnn}
Peter Battaglia, Jessica Blake~Chandler Hamrick, Victor Bapst, Alvaro Sanchez, Vinicius Zambaldi, Mateusz Malinowski, Andrea Tacchetti, David Raposo, Adam Santoro, Ryan Faulkner, Caglar Gulcehre, Francis Song, Andy Ballard, Justin Gilmer, George~E. Dahl, Ashish Vaswani, Kelsey Allen, Charles Nash, Victoria~Jayne Langston, Chris Dyer, Nicolas Heess, Daan Wierstra, Pushmeet Kohli, Matt Botvinick, Oriol Vinyals, Yujia Li, and Razvan Pascanu.
\newblock Relational inductive biases, deep learning, and graph networks.
\newblock \emph{arXiv}, 2018.

\bibitem[Bear et~al.(2021)Bear, Wang, Mrowca, Binder, Tung, RT, Holdaway, Tao, Smith, Sun, Li, Kanwisher, Tenenbaum, Yamins, and Fan]{physion}
Daniel Bear, Elias Wang, Damian Mrowca, Felix Binder, Hsiao-Yu Tung, Pramod RT, Cameron Holdaway, Sirui Tao, Kevin Smith, Fan-Yun Sun, Fei-Fei Li, Nancy Kanwisher, Josh Tenenbaum, Dan Yamins, and Judith Fan.
\newblock Physion: Evaluating physical prediction from vision in humans and machines.
\newblock In \emph{Proceedings of the Neural Information Processing Systems Track on Datasets and Benchmarks}. Curran, 2021.

\bibitem[Boulch(2020)]{boulch2020convpoint}
Alexandre Boulch.
\newblock Convpoint: Continuous convolutions for point cloud processing.
\newblock \emph{Computers \& Graphics}, 88:\penalty0 24--34, 2020.

\bibitem[Cao et~al.(2022)Cao, Chai, Li, and Jiang]{cao2022bi}
Yadi Cao, Menglei Chai, Minchen Li, and Chenfanfu Jiang.
\newblock Bi-stride multi-scale graph neural network for mesh-based physical simulation.
\newblock \emph{arXiv preprint arXiv:2210.02573}, 2022.

\bibitem[Choy et~al.(2019)Choy, Gwak, and Savarese]{choy20194d}
Christopher Choy, JunYoung Gwak, and Silvio Savarese.
\newblock 4d spatio-temporal convnets: Minkowski convolutional neural networks.
\newblock In \emph{Proceedings of the IEEE/CVF conference on computer vision and pattern recognition}, pages 3075--3084, 2019.

\bibitem[Coumans(2015)]{bullet}
Erwin Coumans.
\newblock Bullet physics simulation.
\newblock In \emph{ACM SIGGRAPH 2015 Courses}, New York, NY, USA, 2015. Association for Computing Machinery.

\bibitem[Fortunato et~al.(2022)Fortunato, Pfaff, Wirnsberger, Pritzel, and Battaglia]{fortunato2022multiscale}
Meire Fortunato, Tobias Pfaff, Peter Wirnsberger, Alexander Pritzel, and Peter Battaglia.
\newblock Multiscale meshgraphnets.
\newblock In \emph{ICML 2022 2nd AI for Science Workshop}, 2022.

\bibitem[Gilmer et~al.(2017)Gilmer, Schoenholz, Riley, Vinyals, and Dahl]{msg_passing}
Justin Gilmer, Samuel~S. Schoenholz, Patrick~F. Riley, Oriol Vinyals, and George~E. Dahl.
\newblock Neural message passing for quantum chemistry.
\newblock In \emph{Proceedings of the 34th International Conference on Machine Learning - Volume 70}, page 1263–1272. JMLR.org, 2017.

\bibitem[Graham et~al.(2018)Graham, Engelcke, and Van Der~Maaten]{graham20183d}
Benjamin Graham, Martin Engelcke, and Laurens Van Der~Maaten.
\newblock 3d semantic segmentation with submanifold sparse convolutional networks.
\newblock In \emph{Proceedings of the IEEE conference on computer vision and pattern recognition}, pages 9224--9232, 2018.

\bibitem[Greff et~al.(2022)Greff, Belletti, Beyer, Doersch, Du, Duckworth, Fleet, Gnanapragasam, Golemo, Herrmann, Kipf, Kundu, Lagun, Laradji, Liu, Meyer, Miao, Nowrouzezahrai, Oztireli, Pot, Radwan, Rebain, Sabour, Sajjadi, Sela, Sitzmann, Stone, Sun, Vora, Wang, Wu, Yi, Zhong, and Tagliasacchi]{greff2021kubric}
Klaus Greff, Francois Belletti, Lucas Beyer, Carl Doersch, Yilun Du, Daniel Duckworth, David~J Fleet, Dan Gnanapragasam, Florian Golemo, Charles Herrmann, Thomas Kipf, Abhijit Kundu, Dmitry Lagun, Issam Laradji, Hsueh-Ti~(Derek) Liu, Henning Meyer, Yishu Miao, Derek Nowrouzezahrai, Cengiz Oztireli, Etienne Pot, Noha Radwan, Daniel Rebain, Sara Sabour, Mehdi S.~M. Sajjadi, Matan Sela, Vincent Sitzmann, Austin Stone, Deqing Sun, Suhani Vora, Ziyu Wang, Tianhao Wu, Kwang~Moo Yi, Fangcheng Zhong, and Andrea Tagliasacchi.
\newblock Kubric: a scalable dataset generator.
\newblock 2022.

\bibitem[Grigorev et~al.(2023)Grigorev, Black, and Hilliges]{grigorev2023hood}
Artur Grigorev, Michael~J Black, and Otmar Hilliges.
\newblock Hood: Hierarchical graphs for generalized modelling of clothing dynamics.
\newblock In \emph{Proceedings of the IEEE/CVF Conference on Computer Vision and Pattern Recognition}, pages 16965--16974, 2023.

\bibitem[Han et~al.(2022)Han, Huang, Ma, Li, Tenenbaum, and Gan]{han2022learning}
Jiaqi Han, Wenbing Huang, Hengbo Ma, Jiachen Li, Josh Tenenbaum, and Chuang Gan.
\newblock Learning physical dynamics with subequivariant graph neural networks.
\newblock \emph{Advances in Neural Information Processing Systems}, 35:\penalty0 26256--26268, 2022.

\bibitem[Huber(1992)]{Huber1992}
Peter~J. Huber.
\newblock \emph{Robust Estimation of a Location Parameter}, pages 492--518.
\newblock Springer New York, New York, NY, 1992.

\bibitem[Li et~al.(2023)Li, Wu, Fern, and Fuxin]{li2023improving}
Xingyi Li, Wenxuan Wu, Xiaoli~Z Fern, and Li Fuxin.
\newblock Improving the robustness of point convolution on k-nearest neighbor neighborhoods with a viewpoint-invariant coordinate transform.
\newblock In \emph{Proceedings of the IEEE/CVF Winter Conference on Applications of Computer Vision}, pages 1287--1297, 2023.

\bibitem[Li et~al.(2018)Li, Wu, Tedrake, Tenenbaum, and Torralba]{li2018learning}
Yunzhu Li, Jiajun Wu, Russ Tedrake, Joshua~B Tenenbaum, and Antonio Torralba.
\newblock Learning particle dynamics for manipulating rigid bodies, deformable objects, and fluids.
\newblock In \emph{International Conference on Learning Representations}, 2018.

\bibitem[Park et~al.(2022)Park, Jeong, Cho, and Park]{park2022fast}
Chunghyun Park, Yoonwoo Jeong, Minsu Cho, and Jaesik Park.
\newblock Fast point transformer.
\newblock In \emph{Proceedings of the IEEE/CVF Conference on Computer Vision and Pattern Recognition}, pages 16949--16958, 2022.

\bibitem[Pfaff et~al.(2021)Pfaff, Fortunato, Sanchez-Gonzalez, and Battaglia]{pfaff2021learning}
Tobias Pfaff, Meire Fortunato, Alvaro Sanchez-Gonzalez, and Peter~W. Battaglia.
\newblock Learning mesh-based simulation with graph networks.
\newblock In \emph{International Conference on Learning Representations}, 2021.

\bibitem[Piloto et~al.(2022)Piloto, Weinstein, Battaglia, and Botvinick]{int_physics}
Luis Piloto, Ari Weinstein, Peter Battaglia, and Matthew Botvinick.
\newblock Intuitive physics learning in a deep-learning model inspired by developmental psychology.
\newblock \emph{Nature Human Behaviour}, 6:\penalty0 1--11, 2022.

\bibitem[Qi et~al.(2017{\natexlab{a}})Qi, Su, Mo, and Guibas]{qi2017pointnet}
Charles~R Qi, Hao Su, Kaichun Mo, and Leonidas~J Guibas.
\newblock Pointnet: Deep learning on point sets for 3d classification and segmentation.
\newblock In \emph{Proceedings of the IEEE conference on computer vision and pattern recognition}, pages 652--660, 2017{\natexlab{a}}.

\bibitem[Qi et~al.(2017{\natexlab{b}})Qi, Yi, Su, and Guibas]{qi2017pointnet++}
Charles~Ruizhongtai Qi, Li Yi, Hao Su, and Leonidas~J Guibas.
\newblock Pointnet++: Deep hierarchical feature learning on point sets in a metric space.
\newblock \emph{Advances in neural information processing systems}, 30, 2017{\natexlab{b}}.

\bibitem[Qi et~al.(2021)Qi, Wang, Pathak, Ma, and Malik]{qi2021learning}
Haozhi Qi, Xiaolong Wang, Deepak Pathak, Yi Ma, and Jitendra Malik.
\newblock Learning long-term visual dynamics with region proposal interaction networks.
\newblock In \emph{ICLR}, 2021.

\bibitem[Riochet et~al.(2022)Riochet, Castro, Bernard, Lerer, Fergus, Izard, and Dupoux]{pvoe_benchmark}
Ronan Riochet, Mario~Ynocente Castro, Mathieu Bernard, Adam Lerer, Rob Fergus, Véronique Izard, and Emmanuel Dupoux.
\newblock Intphys 2019: A benchmark for visual intuitive physics understanding.
\newblock \emph{IEEE Transactions on Pattern Analysis and Machine Intelligence}, 44\penalty0 (9):\penalty0 5016--5025, 2022.

\bibitem[Rukhovich et~al.(2022)Rukhovich, Vorontsova, and Konushin]{rukhovich2022fcaf3d}
Danila Rukhovich, Anna Vorontsova, and Anton Konushin.
\newblock Fcaf3d: Fully convolutional anchor-free 3d object detection.
\newblock In \emph{European Conference on Computer Vision}, pages 477--493. Springer, 2022.

\bibitem[Rukhovich et~al.(2023)Rukhovich, Vorontsova, and Konushin]{rukhovich2023tr3d}
Danila Rukhovich, Anna Vorontsova, and Anton Konushin.
\newblock Tr3d: Towards real-time indoor 3d object detection.
\newblock \emph{arXiv preprint arXiv:2302.02858}, 2023.

\bibitem[Sanchez-Gonzalez et~al.(2020)Sanchez-Gonzalez, Godwin, Pfaff, Ying, Leskovec, and Battaglia]{sanchez2020learning}
Alvaro Sanchez-Gonzalez, Jonathan Godwin, Tobias Pfaff, Rex Ying, Jure Leskovec, and Peter Battaglia.
\newblock Learning to simulate complex physics with graph networks.
\newblock In \emph{International conference on machine learning}, pages 8459--8468. PMLR, 2020.

\bibitem[Schult et~al.(2023)Schult, Engelmann, Hermans, Litany, Tang, and Leibe]{Schult2023mask3d}
Jonas Schult, Francis Engelmann, Alexander Hermans, Or Litany, Siyu Tang, and Bastian Leibe.
\newblock {Mask3D: Mask Transformer for 3D Semantic Instance Segmentation}.
\newblock In \emph{Internationl Conference on Robotics and Automation}, 2023.

\bibitem[Thomas et~al.(2019)Thomas, Qi, Deschaud, Marcotegui, Goulette, and Guibas]{thomas2019kpconv}
Hugues Thomas, Charles~R Qi, Jean-Emmanuel Deschaud, Beatriz Marcotegui, Fran{\c{c}}ois Goulette, and Leonidas~J Guibas.
\newblock Kpconv: Flexible and deformable convolution for point clouds.
\newblock In \emph{Proceedings of the IEEE/CVF international conference on computer vision}, pages 6411--6420, 2019.

\bibitem[Thuerey et~al.(2020)Thuerey, Wei\ss{}enow, Prantl, and Hu]{physics_unet}
Nils Thuerey, Konstantin Wei\ss{}enow, Lukas Prantl, and Xiangyu Hu.
\newblock Deep learning methods for reynolds-averaged navier–stokes simulations of airfoil flows.
\newblock \emph{AIAA Journal}, 58\penalty0 (1):\penalty0 25--36, 2020.

\bibitem[Todorov et~al.(2012)Todorov, Erez, and Tassa]{mujoco}
Emanuel Todorov, Tom Erez, and Yuval Tassa.
\newblock Mujoco: A physics engine for model-based control.
\newblock In \emph{2012 IEEE/RSJ International Conference on Intelligent Robots and Systems}, pages 5026--5033, 2012.

\bibitem[Ummenhofer et~al.(2020)Ummenhofer, Prantl, Thuerey, and Koltun]{Ummenhofer2020Lagrangian}
Benjamin Ummenhofer, Lukas Prantl, Nils Thuerey, and Vladlen Koltun.
\newblock Lagrangian fluid simulation with continuous convolutions.
\newblock In \emph{International Conference on Learning Representations}, 2020.

\bibitem[Wang et~al.(2018)Wang, Suo, Ma, Pokrovsky, and Urtasun]{wang2018deep}
Shenlong Wang, Simon Suo, Wei-Chiu Ma, Andrei Pokrovsky, and Raquel Urtasun.
\newblock Deep parametric continuous convolutional neural networks.
\newblock In \emph{Proceedings of the IEEE conference on computer vision and pattern recognition}, pages 2589--2597, 2018.

\bibitem[Watters et~al.(2017)Watters, Zoran, Weber, Battaglia, Pascanu, and Tacchetti]{NIPS2017_8cbd005a}
Nicholas Watters, Daniel Zoran, Theophane Weber, Peter Battaglia, Razvan Pascanu, and Andrea Tacchetti.
\newblock Visual interaction networks: Learning a physics simulator from video.
\newblock In \emph{Advances in Neural Information Processing Systems}. Curran Associates, Inc., 2017.

\bibitem[Wu et~al.(2019)Wu, Qi, and Fuxin]{wu2019pointconv}
Wenxuan Wu, Zhongang Qi, and Li Fuxin.
\newblock Pointconv: Deep convolutional networks on 3d point clouds.
\newblock In \emph{Proceedings of the IEEE/CVF Conference on computer vision and pattern recognition}, pages 9621--9630, 2019.

\bibitem[Wu et~al.(2023)Wu, Fuxin, and Shan]{wu2023pointconvformer}
Wenxuan Wu, Li Fuxin, and Qi Shan.
\newblock Pointconvformer: Revenge of the point-based convolution.
\newblock In \emph{Proceedings of the IEEE/CVF Conference on Computer Vision and Pattern Recognition}, pages 21802--21813, 2023.

\bibitem[Wu et~al.(2022)Wu, Lao, Jiang, Liu, and Zhao]{wu2022point}
Xiaoyang Wu, Yixing Lao, Li Jiang, Xihui Liu, and Hengshuang Zhao.
\newblock Point transformer v2: Grouped vector attention and partition-based pooling.
\newblock \emph{Advances in Neural Information Processing Systems}, 35:\penalty0 33330--33342, 2022.

\bibitem[Xu et~al.(2022)Xu, Lei, Dobriban, and Daniilidis]{xu2022unified}
Yinshuang Xu, Jiahui Lei, Edgar Dobriban, and Kostas Daniilidis.
\newblock Unified fourier-based kernel and nonlinearity design for equivariant networks on homogeneous spaces.
\newblock In \emph{International Conference on Machine Learning}, pages 24596--24614. PMLR, 2022.

\bibitem[Zhang et~al.(2019)Zhang, Hua, Rosen, and Yeung]{zhang2019rotation}
Zhiyuan Zhang, Binh-Son Hua, David~W Rosen, and Sai-Kit Yeung.
\newblock Rotation invariant convolutions for 3d point clouds deep learning.
\newblock In \emph{2019 International conference on 3d vision (3DV)}, pages 204--213. IEEE, 2019.

\bibitem[Zhao et~al.(2021)Zhao, Jiang, Jia, Torr, and Koltun]{zhao2021point}
Hengshuang Zhao, Li Jiang, Jiaya Jia, Philip~HS Torr, and Vladlen Koltun.
\newblock Point transformer.
\newblock In \emph{Proceedings of the IEEE/CVF international conference on computer vision}, pages 16259--16268, 2021.

\end{thebibliography}
}


\maketitlesupplementary

\setcounter{section}{0}

\begin{figure*}[ht!]
\centering
\includegraphics[width=0.8\linewidth]{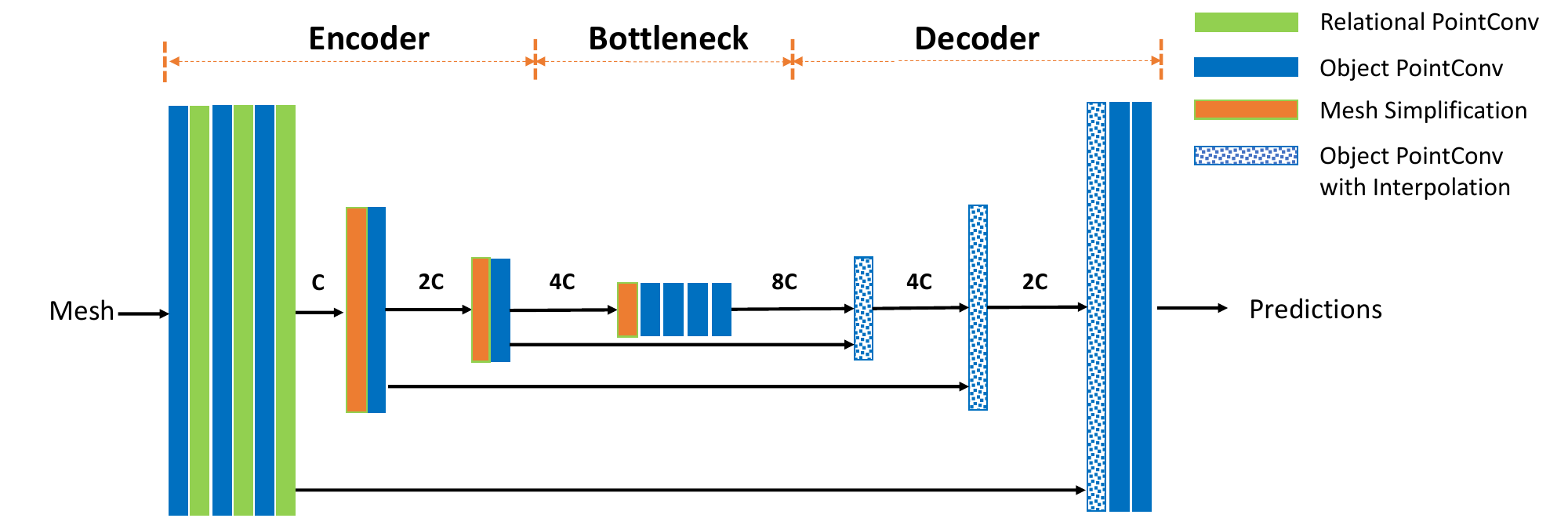}
\caption{The proposed U-Net architecture for mesh inputs. Unlike its counterpart designed for dense point cloud inputs, we only utilize relational PointConv at the highest resolutions, as some object shapes might not be accurately represented with fewer mesh vertices. In this U-net architecture, the downsampling layers only have object PointConv that propagates collision effects over the mesh vertices of each object.}
\label{fig:mesh_unet}
\end{figure*}

\begin{table*}[ht!]
\centering
\vskip 0.15in
\scalebox{0.77}{
\begin{tabular}{lcccccccc}
\toprule
{} & Dominoes & Contain & Link & Support & Drop & Collide & Roll & Average \\
\midrule
3 blocks (1 encoding, 1 bottleneck, 1 decoding) &  $90.2 \pm 1.1$ & $72.2 \pm 3.6$ & $73.1 \pm 4.1$ & $83.3 \pm 0.9$ & $85.8 \pm 2.2$& $91.1 \pm 1.4$ &  $87.5 \pm 0.3$  & 83.3 \\
6 blocks (2 encoding, 2 bottleneck, 2 decoding) & $88.7 \pm 1.6$ & $75.8 \pm 1.7$ & $74.2 \pm 0.8$ & $83.8 \pm 1.7$ & $87.4 \pm 0.9$ & $91.6 \pm 0.6$ & $87.3 \pm 0.0$ & $84.1$  \\
9 blocks (3 encoding, 3 bottleneck, 3 decoding) & $90.2 \pm 1.3$ & $75.1 \pm 2.1$ &  $75.3 \pm 1.9$ & $83.6 \pm 3.5$ & $88.0 \pm 0.9$ & $90.9 \pm 0.8$ & $87.5 \pm 0.3$ & $84.4$ \\
\bottomrule
\end{tabular}
}
\caption{Ablation on different numbers of interaction blocks on the Physion dataset with the contact prediction accuracy (\%)} \label{ablation_layer_num}
\end{table*}

\begin{table*}[ht!]
\centering
\scalebox{0.77}{
\begin{tabular}{lcccccccc}
\toprule
{} & Dominoes & Contain & Link & Support & Drop & Collide & Roll & Average\\
\midrule
DPI \cite{li2018learning} & $82.3 \pm 1.3$  & $72.3 \pm 1.8$ & $63.7 \pm 2.2$ & $64.8 \pm 2.0$ & $70.7 \pm 0.8$ & $84.4 \pm 0.7$ & $82.3 \pm 0.6$    & $74.4$ \\
\midrule
PointConv U-Net w/o object-centric PointConv  & $\mathbf{88.2} \pm 0.3$ & $71.8 \pm 3.5$ & $70.0 \pm 4.3$ & $\mathbf{82.4} \pm 2.5$ & $82.7 \pm 1.5$ & $\mathbf{90.9} \pm 0.3$ & $\mathbf{87.8} \pm 0.3$ & $\mathbf{82.0}$\\
PointConv U-Net  & $\mathbf{90.2} \pm 1.3$ & $75.1 \pm 2.1$ &  $75.3 \pm 1.9$ & $\mathbf{83.6} \pm 3.5$ & $\mathbf{88.0} \pm 0.9$ & $\mathbf{90.9} \pm 0.8$ & $\mathbf{87.5} \pm 0.3$ & $\mathbf{84.4}$ \\
\bottomrule
\end{tabular}
}
\caption{Ablation on the interaction block design that creates the separation between object and relational PointConv} \label{no_relational_pointconv}
\end{table*}

In this document, we provide more details about our method, which is presented in the main paper, as well as additional experimental results.

\section{Finding Interaction Points}

Here, we describe an approach used for identifying interaction points between mesh faces of different objects. Initially, we sample points from mesh faces uniformly and calculate the distances between these sampled points across the mesh faces of different objects. Note that a pair of mesh faces can have multiple pairs of sampled points, given that each mesh face can keep multiple sampled points. Interaction point pairs are identified when the distance between sampled points on different faces falls below a specified threshold. Subsequently, any pair of mesh faces containing at least one interaction point pair is considered to have a distance approximately within the threshold. Note that our proposed Relational PointConv, designed for processing mesh data directly, is also compatible with other approaches for identifying closely positioned mesh faces, such as the Bounding Volume Hierarchy algorithm used in \cite{allen2022learning}.

\section{Speed Comparison}

We ran SGNN's \cite{han2022learning} and our inference code with the same point cloud input on the same machine with RTX 3090. Averaging across the Physion scene types, our inference code ran 17.10 fps while SGNN's code ran 38.9 fps. The SGNN code is built based on the DPI code, so we expect DPI to run at a similar speed. Our implementation is still fast, given DPI and SGNN are shallow networks with only two layers of hierarchy.

\section{Point Cloud Generation}
For both datasets, ground truth meshes are available. Hence, we uniformly sample points from the mesh faces to obtain a point cloud dense on object surfaces. Then, we employ an initial grid sampling step with a voxel size of $0.05$ for Physion and $0.2$ for Kubric. This process subsamples the point cloud with one point per voxel. Afterward, within the U-Net architecture, we utilize different voxel sizes for downsampling at different levels. For Physion, we employ voxel sizes of $0.075$, $0.1125$, and $0.16875$ at different downsampling levels. For Kubric, we use voxel sizes of $0.3$, $0.45$, and $0.675$ at different  downsampling levels. 

For the distance threshold $r$ used in KNN for relational PointConv, we select $0.1$ as the threshold at the highest resolution in U-Net for Physion. Subsequently, for the downsampling levels, we employ thresholds of $0.15$, $0.225$, and $0.3375$. Regarding Kubric, we utilize a threshold of $0.4$ at the highest resolution, with thresholds of $0.6$, $0.9$, and $1.35$ for the downsampling levels.

\section{Mesh Preprocessing}
We use the Kubric dataset for our mesh experiments. Since the original meshes in the Kubric dataset come with too many vertices, we first perform mesh simplification via vertex clustering with a voxel size of $0.2$ and $0.4$ for Movi-A and Movi-C, respectively. For the downsampling layers in the U-Net architecture, we adopt the mesh simplification method proposed in \cite{grigorev2023hood} instead of the grid downsampling used for point clouds.

\section{Interaction PointConv U-Net for Mesh}

We used a variant of the PointConv U-Net architecture shown in Fig. \ref{fig:mesh_unet} for our mesh experiments. This choice was made because applying relational PointConv to downsampled meshes may result in inaccurate face-to-face collision modeling unless the downsampling process accurately preserves the shapes of object surfaces.

\section{Training details}
We utilized the Adam optimizer with an initial learning rate of $0.001$ and trained the model for 30 epochs for the Physion dataset. For the Kubric dataset, we employed an initial learning rate of $0.005$ and trained the model for 15 epochs. During each training iteration, a single input frame was randomly selected from each training video. We defined one epoch as the point when every video in the training dataset is sampled $N$ times for input frames. In the case of Physion, we set $N$ to $8$, consistent with previous studies \cite{han2022learning, physion}, while for Kubric, we set $N$ to $2$.

\section{Additional Ablation Results}

In this section, we present results from additional ablation experiments that we conduct with acceleration prediction models. In Table \ref{ablation_layer_num}, we compare the performance using different numbers of interaction blocks used in the U-Net. For this experiment, we use the same number of interaction blocks in the encoder, bottleneck, and decoder of U-Net for simplicity, but note that a different number of blocks can be used for the bottleneck in practice depending on the needs (e.g., more interaction blocks if longer-range force propagation needs to be modeled). The result with $3$ blocks corresponds to the scenario where a single downsampling and upsampling step is employed. One can see that results for most scenarios in the Physion dataset do not change significantly with further downsampling.

In Table \ref{no_relational_pointconv}, we compare our approach with an alternative where the entire point cloud is directly processed without separate object and relational PointConv layers. Note that this baseline requires an expensive KNN neighbor search every frame across all PointConv layers, whereas the neighbor search needs to be done only once when the object PointConv layers are used for rigid body objects. Results in Table~\ref{no_relational_pointconv} show that the non-object-centric baseline can still perform well, especially for scenes such as Roll and Collide, where objects are well separated. This result makes sense, as when objects are well separated most of the time, the KNN neighbors generally come from the same object except for a few collision events.  Also, note that our non-object-centric baseline generally performs better than other non-object-centric baselines such as DPI \cite{li2018learning}.

\end{document}